\documentclass{egpubl}
\usepackage[dvipsnames]{xcolor}
\usepackage{eg2023}
\usepackage{optidef} 
\usepackage{tikz}
\usepackage{amssymb}
\usepackage{caption}
\usepackage{subcaption}
\usepackage{tikzsymbols}
\usepackage{adjustbox}
\usepackage{bm}
\usetikzlibrary{fit, positioning,matrix, patterns,arrows.meta}
\usepackage{caption,subcaption}

\usepackage{booktabs}

\newcommand{\colvec}[2][.8]{  \scalebox{#1}{    \renewcommand{\arraystretch}{.8}    $\begin{bmatrix}#2\end{bmatrix}$  }
}

\usepackage{xcolor}
\newcommand{\etal}{\textit{et al.}}
\usepackage[T1]{fontenc}
\usepackage{dfadobe}  

\usepackage{cite}  \BibtexOrBiblatex
\electronicVersion
\PrintedOrElectronic

\title[Non-Separable Multi-Dimensional Network Flows for Visual Computing]      {Non-Separable Multi-Dimensional \\Network Flows for Visual Computing}

\author[V. Ehm, D. Cremers, F. Bernard]
{\parbox{\textwidth}{\centering V. Ehm
$^{1,2}$
        D. Cremers$^{1,2}$
                F. Bernard$^3$
                }
        \\
{
\parbox{\textwidth}{\centering $^1$TU Munich   $^2$ Munich Center for Machine Learning $^3$University of Bonn
       }
}
}

\begin{document}

\maketitle
\begin{abstract}
Flows in networks (or graphs) play a significant role in numerous computer vision tasks.
      The scalar-valued edges in these graphs  
often lead to a loss of information and thereby to limitations in terms of expressiveness.
For example, oftentimes high-dimensional data (e.g.~feature descriptors) are mapped to a single scalar value (e.g.~the similarity between two feature descriptors).
   To overcome this limitation, we propose a novel formalism for non-separable multi-dimensional network flows. By doing so, we enable an automatic and adaptive feature selection strategy -- since the flow is defined on a per-dimension basis, the maximizing flow automatically chooses the best matching feature dimensions.
As a proof of concept, we apply our formalism to the multi-object tracking problem and demonstrate that our approach
outperforms scalar formulations on the MOT16 benchmark in terms of robustness to noise.

\ccsdesc[300]{Theory of computation~Design and analysis of algorithms}
\ccsdesc[100]{Theory of computation~Theory and algorithms for application domains}

\printccsdesc   
\end{abstract}  
\section{Introduction}

Network flow algorithms are popular in computer vision and image analysis due to their broad range of applications, including image segmentation~\cite{eriksson2006image} 
or multi-object tracking (MOT)~\cite{zhang2008global}.
Yet, they require that multi-dimensional information is mapped to scalar values which often has the downside that information is lost and thereby the expressiveness limited. 
Hence, in this work we take the first step to explore
 the novel direction of non-separable multi-dimensional flows:
 \begin{itemize}
     \item For the first time we present a \textbf{non-separable multi-commodity flow formulation}, where a flow unit with all its commodities cannot be separated throughout the graph.
     \item As a proof of concept, we show how our formulation can be applied in the context of \textbf{multi-object tracking}, for which we demonstrate 
               that it
          increases the \textbf{robustness to noise}.
      \end{itemize}
 \section{Related Work}
 Garg \etal~\cite{garg1996approximate} extend traditional maximum flows~\cite{harris1955fundamentals} to multi-commodity flows by considering multi-dimensional flows per edge.
Li \etal~\cite{li2010ant} use a binary variable to ensure that only one path per commodity can be used to reach the sink. To the best of our knowledge, there does not exist a non-separable maximum multi-commodity formulation, where non-separable, in this case, means that only one path for all commodities is used.

While traditional MOT approaches~\cite{zhang2008global} minimize the cost through the graph we maximize
a multi-commodity flow.

\section{Method}

\begin{figure}[h!]
\centering
    \begin{tikzpicture}[scale = 0.75]  
        \node[shape=circle,draw=black] (A) at (0,0.5) {s};
        \node[shape=circle,draw=black] (B) at (3,1) {1};
        \node[shape=circle,draw=black] (C) at (3,0) {2};
        \node[shape=circle,draw=black] (D) at (6,0.5) {t};

        \path [->] (A) edge node[above] {$\textcolor{blue}{2} / \textcolor{red}{2}$} (B);
        \path [->](A) edge node[below] {$\textcolor{blue}{3} / \textcolor{red}{5}$} (C);
        \path [->](B) edge node[above] {$\textcolor{blue}{2} / \textcolor{red}{4}$} (D);
        \path [->](C) edge node[below] {$\textcolor{blue}{3} / \textcolor{red}{3}$} (D);
    \end{tikzpicture}
   \caption{\textbf{Scalar Maximum Flow:} Maximize the flow (blue) from source node $s$ to sink node $t$ while not exceeding the capacity (red).} 
   \label{fig:scalar_max}
\end{figure}
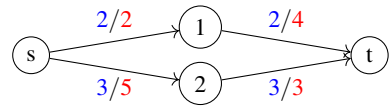

A graph is a tuple $\mathcal{G}=(\mathcal{V},\mathcal{E})$,
where $\mathcal{V}$ represents the nodes and $\mathcal{E} \subseteq \mathcal{V} \times \mathcal{V}$ the set of edges. 
In the scalar maximum flow approach~\cite{harris1955fundamentals} the sum of flows $f_{uv}\in \mathbb{R}$ from the source node $s \in \mathcal{V}$ to the sink node $t \in \mathcal{V}$ is maximized while not exceeding the capacity $c_{uv}\in \mathbb{R}_+$ on every edge (see Figure~\ref{fig:scalar_max}).

In comparison to the scalar approach, we assign a 
capacity vector $c_{uv} \in \mathbb{R}_+^k$ with $k$ 
dimensions to every edge $(u,v) \in \mathcal{E}$.
Additionally, we define a multi-dimensional flow 
$f_{uv} \in \mathbb{R}^k$ 
and a a decision variable $b_{uv} \in \{0,1\}$, indicating whether an edge is active or not. 
An active edge means that the flow can be non-zero, whereas an inactive edge ensures that its flow is the zero vector.
We impose the capacity constraint that each flow $f_{uv}$ through an edge must be elementwise smaller than its capacity.
The sum of flows of all incoming edges at every node needs to equal the sum of flows of all outgoing edges (flow conservation).
Further, we constrain that for each node only a single incoming and a single outgoing flow vector may have a non-zero flow, thereby ensuring that incoming flow cannot be separated into multiple edges (node count constraint).
To fix the total number of flow entities leaving the source node, and entering the target node to a constant $d$ we define the total count constraint.
Overall, the resulting problem is a mixed-integer programming (MIP) problem that reads
\begin{maxi*}|l|
    {\stackrel{f_{uv} \in \mathbb{R}^k}{b_{uv} \in \{0,1\}}} {\sum \limits_{v:s \to v} f_{sv}^T \bold{1}}{}{}    \addConstraint{f_{uv}}{\leq b_{uv}c_{uv}, \quad}{\forall (u,v) \in E   \text{ (capacity)}}
    \addConstraint{ \sum \limits_{u:u \to v} f_{uv}}{= \sum \limits_{w:v \to w} f_{vw},\quad}{\forall v \neq s,t  \text{ (flow cons.)}} 
    \addConstraint{ \sum \limits_{u:u \to v} b_{uv}}{= \sum \limits_{w:v \to w} b_{vw} = 1,\quad}{\forall v \neq s,t  \text{ (node count)}} 
    \addConstraint{ \sum \limits_{u:s \to u} b_{su}}{= \sum \limits_{v:v \to t} b_{vt} = d,\quad}{\forall u,v \neq s,t  \text{ (total count)}} 
    \addConstraint{f_{uv}}{\geq \textbf{0}}{\forall (u,v) \in E  \text{ (non-neg.)}.}
    \label{eq:MIP_ours}
\end{maxi*}

\subsection{Application to Multi-Object Tracking}\label{sec:application-to-muktiobject-tracking}

\begin{figure}[h]
    \centering
        \begin{tikzpicture}[scale = 0.75] 
        \def\A{20};
        \def\B{2};

        \draw[draw=black] (0,7) rectangle++ (2,1);

        \draw[draw=black] (4,7) rectangle++ (2,1);

        \draw[draw=black] (8,7) rectangle++ (2,1);

        \node at (1,7.66) {\textcolor{blue}{\Strichmaxerl[2]}};
        \node at (5,7.66) {\textcolor{blue}{\Strichmaxerl[2]}};
        \node at (9,7.66) {\textcolor{blue}{\Strichmaxerl[2]}};

        \node at (1.3,7.3) {\textcolor{red}{\Strichmaxerl[2][90+\A*18][-\A*18][-45+\A*5][45-\A*5]}};
        \node at (5.7,7.3) {\textcolor{red}{\Strichmaxerl[2][90+\A*18][-\A*18][-45+\A*5][45-\A*5]}};

        \node at (4.2,7.3) {\textcolor{ForestGreen}{\Strichmaxerl[2][90+\B*18][-\B*18][-45+\B*5][45-\B*5]}};
        \node at (8.5,7.3) {\textcolor{ForestGreen}{\Strichmaxerl[2][90+\B*18][-\B*18][-45+\B*5][45-\B*5]}};

        \draw[draw=black] (0,5.7) rectangle++ (2,1);
        \node at (0.7,5.4) {Frame $t$};

        \draw[draw=black] (4,5.7) rectangle++ (2,1);
        \node at (5,5.4) {Frame $t+1$};

        \draw[draw=black] (8,5.7) rectangle++ (2,1);
        \node at (9,5.4) {Frame $t+2$};

        \node at (1,6.35) {$\textcolor{blue}{\colvec{1\\0}}$};
        \node at (5,6.35) {$\textcolor{blue}{\colvec{1\\0}}$};
        \node at (9,6.35) {$\textcolor{blue}{\colvec{1\\0}}$};

        \node at (1.4,6.1) {$\textcolor{red}{\colvec{0\\1}}$};
        \node at (5.7,6.1) {$\textcolor{red}{\colvec{0\\1}}$};

        \node at (4.4,6.1) {$\textcolor{ForestGreen}{\colvec{0.5\\0.5}}$};
        \node at (8.45,6.1) {$\textcolor{ForestGreen}{\colvec{0.5\\0.5}}$};

    \end{tikzpicture}
        \begin{tikzpicture}[scale = 0.75]
        \def\opac{0.5};
        \pgfdeclarelayer{bg}    
        \pgfsetlayers{bg,main}  
        \node[shape=circle,draw=black, fill=white, fill opacity=\opac, text opacity=1] (s) at (5,5.2) {$s$};
        \node[shape=circle,draw=black, fill=white, fill opacity=\opac, text opacity=1] (t) at (5,1) {$t$};
        \node[shape=circle,draw=black, fill=white, fill opacity=\opac, text opacity=1] (u1) at (0,3.5) {$e_1$};
        \node[shape=circle,draw=black, fill=white, fill opacity=\opac, text opacity=1] (u2) at (0,2.5) {$e_2$};
        \node[shape=circle,draw=black, fill=white, fill opacity=\opac, text opacity=1] (v1) at (2,3.5) {$o_1$};
        \node[shape=circle,draw=black, fill=white, fill opacity=\opac, text opacity=1] (v2) at (2,2.5) {$o_2$};

        \node[shape=circle,draw=black, fill=white, fill opacity=\opac, text opacity=1] (u3) at (4,4) {$e_3$};
        \node[shape=circle,draw=black, fill=white, fill opacity=\opac, text opacity=1] (u4) at (4,3) {$e_4$};
        \node[shape=circle,draw=black, fill=white, fill opacity=\opac, text opacity=1] (u5) at (4,2) {$e_5$};
        \node[shape=circle,draw=black, fill=white, fill opacity=\opac, text opacity=1] (v3) at (6,4) {$o_3$};
        \node[shape=circle,draw=black, fill=white, fill opacity=\opac, text opacity=1] (v4) at (6,3) {$o_4$};
        \node[shape=circle,draw=black, fill=white, fill opacity=\opac, text opacity=1] (v5) at (6,2) {$o_5$};

        \node[shape=circle,draw=black, fill=white, fill opacity=\opac, text opacity=1] (u6) at (8,3.5) {$e_6$};
        \node[shape=circle,draw=black, fill=white, fill opacity=\opac, text opacity=1] (u7) at (8,2.5) {$e_7$};
        \node[shape=circle,draw=black, fill=white, fill opacity=\opac, text opacity=1] (v6) at (10,3.5) {$o_6$};
        \node[shape=circle,draw=black, fill=white, fill opacity=\opac, text opacity=1] (v7) at (10,2.5) {$o_7$};
        \path [->] (u1) edge[draw=blue] node[above, fill = white,  opacity=0.8, text opacity=1] {$\textcolor{blue}{\colvec{1\\0}}$} (v1);
    \begin{pgfonlayer}{bg}    

        \path [->] (u2) edge[draw=red] node[above] {$\textcolor{red}{\colvec{0\\1}}$} (v2);
        \path [->] (u3) edge[draw=blue] node[above] {$\textcolor{blue}{\colvec{1\\0}}$} (v3);
        \path [->] (u4) edge[draw=red] node[above] {$\textcolor{red}{\colvec{0\\1}}$} (v4);
        \path [->] (u5) edge[draw=ForestGreen] node[above] {$\textcolor{ForestGreen}{\colvec{0.5\\0.5}}$} (v5);
        \path [->] (u6) edge[draw=blue] node[above] {$\textcolor{blue}{\colvec{1\\0}}$} (v6);
        \path [->] (u7) edge[draw=ForestGreen] node[above] {$\textcolor{ForestGreen}{\colvec{0.5\\0.5}}$} (v7);

        \path [->] (v1) edge[draw=blue] node[left] {} (u3);
        \path [->] (v2) edge[draw=gray] node[left] {} (u3);
        \path [->] (v1) edge[draw=gray] node[left] {} (u4);
        \path [->] (v2) edge[draw=red] node[left] {} (u4);
        \path [->] (v1) edge[draw=gray] node[left] {} (u5);
        \path [->] (v2) edge[draw=gray] node[left] {} (u5);

        \path [->] (v3) edge[draw=blue] node[left] {} (u6);
        \path [->] (v3) edge[draw=gray] node[left] {} (u7);
        \path [->] (v4) edge[draw=gray] node[left] {} (u6);
        \path [->] (v4) edge[draw=gray] node[left] {} (u7);
        \path [->] (v5) edge[draw=gray] node[left] {} (u6);
        \path [->] (v5) edge[draw=ForestGreen] node[left] {} (u7);

        \path [->] (s) edge[draw=blue, bend right] node[left] {} (u1);
        \path [->] (s) edge[draw=red, bend right] node[left] {} (u2);
        \path [->] (s) edge[draw=gray, bend right] node[left] {} (u3);
        \path [->] (s) edge[draw=gray, bend right] node[left] {} (u4);
        \path [->] (s) edge[draw=ForestGreen, bend right] node[left] {} (u5);
        \path [->] (s) edge[draw=gray, bend left] node[left] {} (u6);
        \path [->] (s) edge[draw=gray,  bend left] node[left] {} (u7);

        \path [->] (v2) edge[draw=gray, bend right] node[left] {} (t);
        \path [->] (v3) edge[draw=gray, bend left] node[left] {} (t);
        \path [->] (v4) edge[draw=red, bend left] node[left] {} (t);
        \path [->] (v5) edge[draw=gray, bend left] node[left] {} (t);
        \path [->] (v6) edge[draw=blue, bend left] node[left] {} (t);
        \path [->] (v7) edge[draw=ForestGreen, bend left] node[left] {} (t);
    \end{pgfonlayer}
    \end{tikzpicture}
    \caption{
    \textbf{Graph construction:} Objects of the 
        sample frames (top row) are represented by feature vectors (middle row).
    These vectors are set as capacity on the corresponding object edges     (bottom row).
                        }
\label{fig:encoding}
\end{figure}
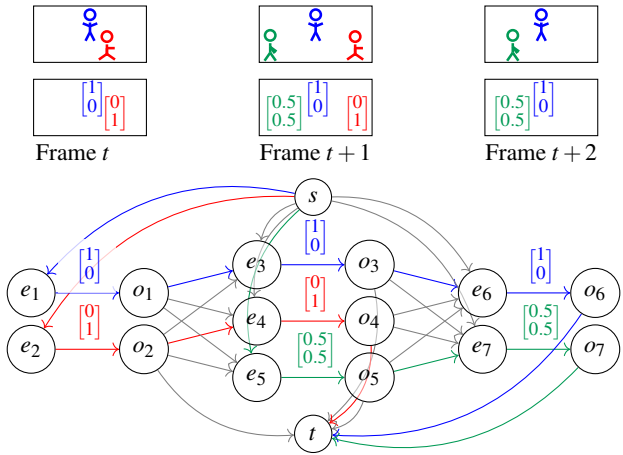

Similar to the 
scalar flow model for MOT~\cite{zhang2008global}, we define three different types of edges: (i) observation edges, (ii) transition edges, and (iii) enter/exit edges.
We represent every detected object $x_i$ in a frame by a vector and assign this vector to an observation edge $(e_i, o_i)$ as a capacity vector.
We show an example graph in Figure~\ref{fig:encoding}.

By connecting the source to all nodes $e_i$ and connecting all nodes $o_i$ to the sink, objects can appear/disappear at any time (enter/exit nodes).
Additionally, we connect all end nodes $o_i$ with all start nodes $e_j$ in a timespan $\Delta t$ (transition edges), such that trajectories can skip 
the next $\Delta t$ frames.
Enter, exit, and transition edges are assigned infinite capacity. 
While the graph connections look similar to the scalar 
method~\cite{zhang2008global}, the capacities are set differently such that we can send vector-valued 
instead of scalar-valued flows.

\section{Results}
We choose training sequences (2,4,5,9,10 and 11) of the MOT16 Challenge dataset~\cite{milan2016mot16} as a benchmark.
We provide the ground truth boxes and the ground truth number of individual objects to the algorithms such that we focus on the tracking rather than the detection part.
For our experiments, we use two different feature descriptors: color histograms 
and deep features(\cite{he2016deep}). To evaluate the robustness, 
we add random Gaussian noise with different variances to every image.
We reduce the runtime by pruning and batch splitting.
To evaluate our algorithm we use a metric that normalizes the identity switches (IDSW) by the total number of ground truth boxes (GT) for all frames $t$: $IDSW_{norm} = \frac{\sum_t IDSW_t}{\sum_t GT_t}$.

\newcommand{\noiseFigMOTA}[2]{
    \begin{subfigure}[b]{.2\textwidth}
        \centering
        \small{\quad \quad #2}
        \includegraphics[width=1\textwidth]{./images/mota_noise_#1_new_2.pdf}
    \end{subfigure}
    }
In Figure~\ref{fig:noise_mota}, we show that our algorithm performs substantially better on noisy data than the scalar baseline method~\cite{zhang2008global} when using color features and deep features.
Our method automatically allows to select (per object) feature dimensions that have the smallest variability across the entire sequence.
The scalar method is not able to dynamically select features, since it computes a scalar score that summarizes feature similarities, and thus performs worse.

\begin{figure}[h!]
    \centering
        \noiseFigMOTA{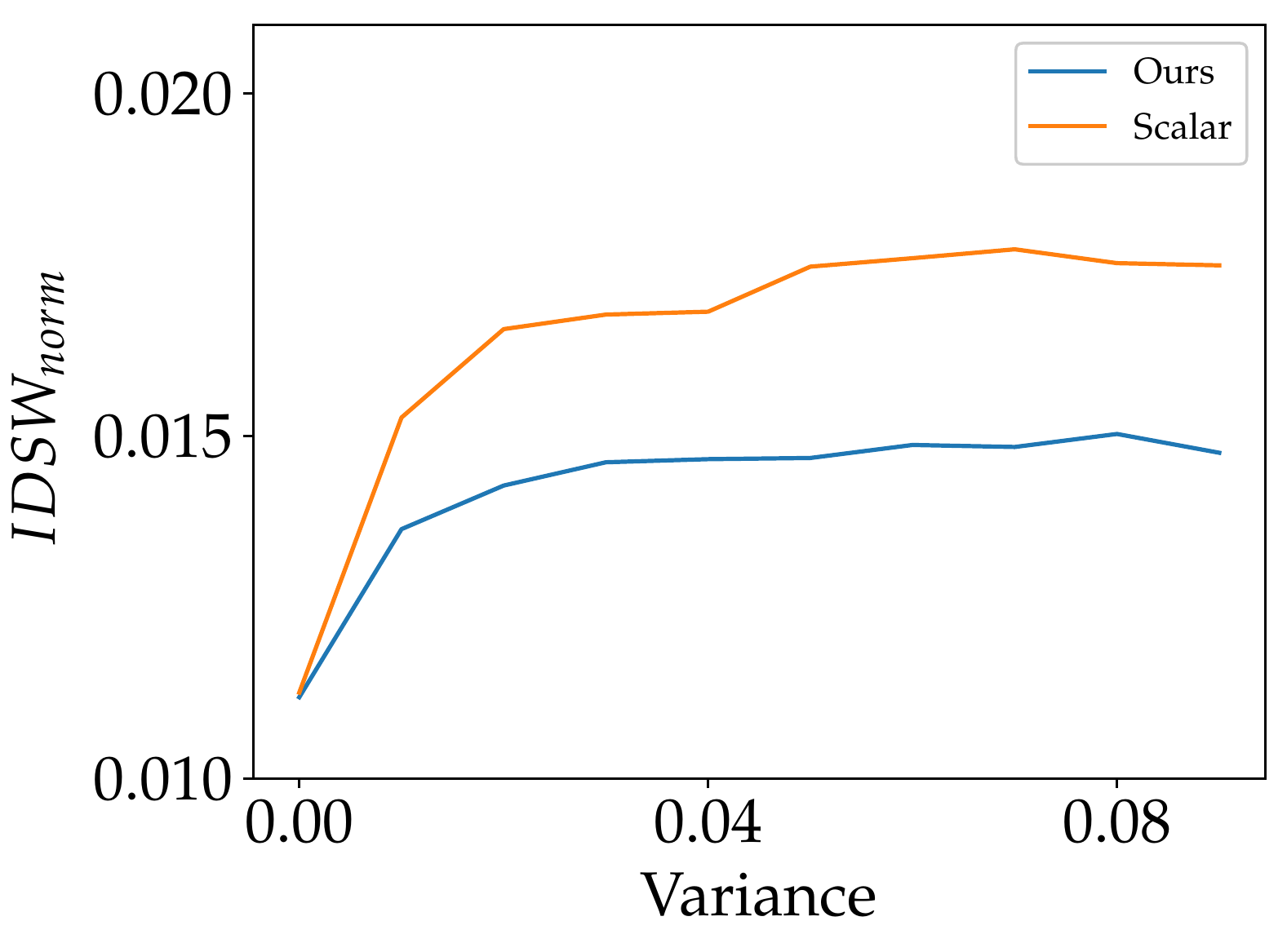}{Color}
        \noiseFigMOTA{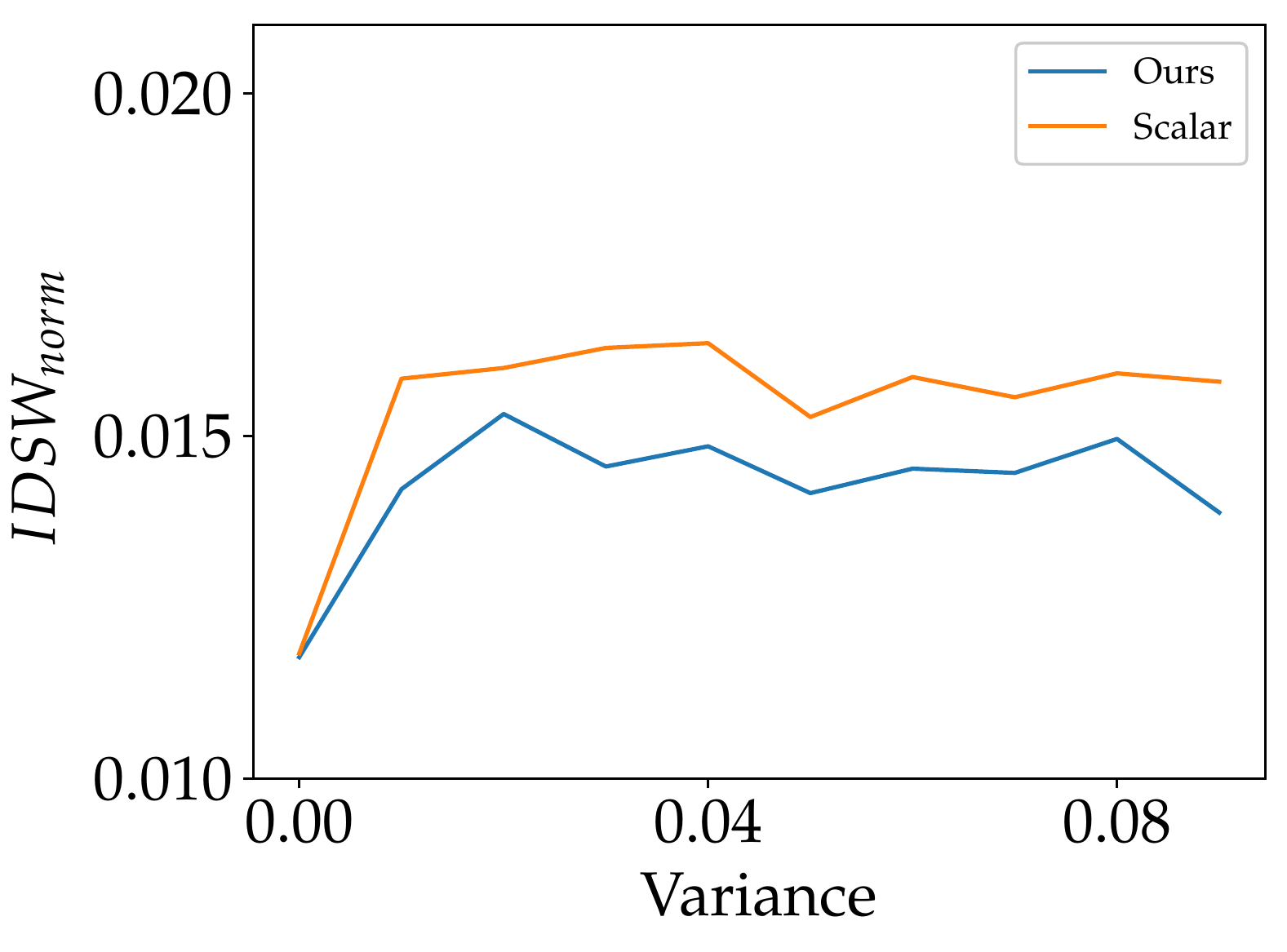}{Deep}
\caption{\textbf{$\mathbf{IDSW_{norm}}$ ($\downarrow$) for different noise values}: Our method performs better on noisy images than the scalar method with different feature descriptors (left: color, right: deep features).
        }
\label{fig:noise_mota}
\end{figure}

\section{Conclusion}
For the first time we conceptualized a non-separable multi-dimensional maximum flow formulation, and
we demonstrated that such a formalism can naturally be applied to multi-object tracking. Since our flow is defined on a per-dimension basis, the maximizing flow automatically chooses the subset of feature dimensions that best match across a sequence.

\bibliographystyle{eg-alpha-doi} 
\bibliography{egbibsample}

\end{document}